# Dual flow fusion model for concrete surface crack segmentation

Yuwei Duan

Corresponding author(s). E-mail(s): zy2006211@buaa.edu.cn

**Abstract** The existence of cracks and other damages pose a significant threat to the safe operation of transportation infrastructure. Traditional manual detection and ultrasound equipment testing consume a lot of time and resources. With the development of deep learning technology, many deep learning models have been widely applied to practical visual segmentation tasks. The detection method based on deep learning models has the advantages of high detection accuracy, fast detection speed, and simple operation. However, deep learning-based crack segmentation models are sensitive to background noise, have rough edges, and lack robustness. Therefore, this paper proposes a crack segmentation model based on the fusion of dual streams. The image is inputted simultaneously into two designed processing streams to independently extract long-distance dependence and local detail features. The adaptive prediction is achieved through the dual-headed mechanism. Meanwhile, a novel interaction fusion mechanism is proposed to guide the complementary of different feature layers to achieve crack location and recognition in complex backgrounds. Finally, an edge optimization method is proposed to improve the accuracy of segmentation. Experiments show that the F1 value of segmentation results on the DeepCrack[1] public dataset is 93.7% and the IOU value is 86.6%. The F1 value of segmentation results on the CRACK500[2] dataset is 78.1%, and the IOU value is 66.0%.

## I. INTRODUCTION

Railway transportation is the lifeblood of China's national economy. With the increase of operation time and the impact of climate, environment, and service time, cracks and other deterioration phenomena gradually appear on the surface of bridge concrete, posing hidden risks to safe operation. Bridge cracks pose a great threat to the safety of the bridge: accelerating concrete carbonation, reducing the corrosion resistance of various aggressive media, and affecting the structural strength and stability of concrete structures. Therefore, it is necessary to conduct research on intelligent detection systems and key technologies for shallow surface concrete and steel structure cracks in high-speed rail infrastructure. The development of crack detection technology mainly includes traditional image processing, traditional machine learning methods, and deep learning methods. Traditional image processing requires high image preprocessing technology and the detection results are easily affected by factors such as lighting and noise. The machine learning-based detection methods in the early stage used morphological detection methods; later, seed points, tensor voting, and other methods were used, which are computationally complex with low model efficiency and cannot cope with complex scenes.

The deep learning crack detection methods are usually based on convolutional neural networks (CNNs). CNNs have achieved excellent results in multiple visual tasks due to their nonlinearity and rotation invariance characteristics. Convolutional models use convolution operations on images to obtain image features. However, there are limitations, including insufficient perception field and loss of detail information during downsampling. Although existing research has proposed various ways to expand the perception field of CNNs, such as FPN[3] and ASPP[4], the former uses a feature pyramid to enlarge the perception field layer by layer, and the latter uses dilated convolution to increase the perception field, but these methods cannot effectively solve the problem of lack of global features. In crack segmentation tasks, global features provide crack region locations, ensure topological integrity, and help the model identify different texture features in complex backgrounds. To solve the problem of lacking global features, a vision transformer[5] model based on multi-head self-attention calculation was used. The model encodes the image into multiple patches, and calculates self-attention between patches to extract long-distance dependence. However, this calculation method requires much more computation than CNNs, and ignores the relationship between the pixels inside the patch. These details are essential for crack edge perception and are indispensable elements in dense prediction.

To balance global and local information and improve the accuracy of segmentation, some methods have proposed combining CNNs with Transformers. The CvT[6] converts linear projection in self-attention blocks to convolution projection. The GLTB module in EHT uses numerical addition to combine local and global features. The Conformer model relies on Feature Coupling Units (FCUs) to interactively fuse local feature representations and global feature representations at different resolutions. DS-Net[7] proposes a dual-stream framework that cross-fuses convolution and self-attention, with each form of scaling learning aligned with other forms. However, convolution and attention have inherently conflicting properties, which may lead to ambiguity during training

This paper adopts a parallel dual-stream architecture to extract image features, and the parallel streams ensure the independence and integrity of feature extraction. In addition, only feature extraction is not enough. To make global features and local features complement each other, global features obtain probability maps at different levels through multi-head attention, which guides the model to focus on possible areas of interest. Local features, guided by global features, can eliminate noise interference, provide detailed information between pixels, and accurately identify the edges, such as capturing pixel mutation information. Existing fusion methods also enrich features by stacking channels, and the model can adaptively learn features at different levels. Although the above fusion methods have achieved certain results, they overlook effective communication between features. Therefore, we propose a dual-stream feature extraction module and insert a feature interaction process between the two parallel streams, which further enriches semantic features through multi-scale feature fusion. The contributions of this paper are as follows:

1. Using the dual flow structure, the CNN and Transformer branches can respectively preserve the local features and global representation to the maximum extent.

2. Introducing dynamic feature interaction module that employ both global and local features derive from dual stream.

3. Propose an edge optimization strategy based on decoupling to further improve the accuracy of edge information.

## II. RELATED WORKS

With the continuous development of deep learning, deep learning methods have advantages over traditional digital image processing methods in terms of high detection accuracy and good robustness. There are several typical research studies on concrete shallow surface crack detection based on image segmentation: segmentation models based on convolution, segmentation models based on Transformer, and segmentation models combining convolution and Transformer.

**Segmentation based on convolution**. Traditional convolution utilizes an aggregation function on a local receptive field based on the convolution weights, which are shared throughout the entire feature map. This inherent feature brings crucial inductive bias to image processing. Due to the small proportion of pixels occupied by cracks and the similarity of narrow features to edge segmentation in images, the DeepCrack method performs crack segmentation based on the HED[8] edge segmentation network. The DeepCrack model upsamples the results of each downsampling operation to the original size of the image through a bypass branch and then performs loss calculation via deep supervision to further enhance edge accuracy. Wang et al. employed a fully convolutional neural network to detect image cracks and built a CrackFCN[9] more suitable for crack detection. Since cracks occupy only a small proportion of pixels in the image, the CrackFCN cancels the Dropout function in the FCN to reduce local information loss, and uses higher-scale deconvolution layers to expand local details while deepening the FCN's network depth. Experimental results show that this Crack FCN has stronger fine-grained discrimination ability, with higher crack detection accuracy and lower false alarm rate. Wang et al. proposed a new model based on the improved original DeeplabV2[10] to adapt to the special nature of crack damage detection. Compared with previous methods, this method can achieve highly accurate output, and proposes a new method for marking cracks that is beneficial for measuring crack length and width. Yang Min constructed an SPPNet[11] network for tunnel crack detection. SPPNet draws on the ideas of Deeplab and PSPNet[12] to solve this problem, and fuses multi-scale features from different layers, enabling the presence of many higher-level information in the upsampled feature map.

**Segmentation based on Transformer**. The self-attention module employs a weighted average operation based on the input feature context, wherein the attention weights are dynamically calculated through a similarity function between relevant pixel pairs. This flexibility allows the attention module to adaptively focus on different regions and capture more features. The ViT model divides the image into multiple patches, maps them to a linear embedding sequence, encodes them with an encoder, and establishes long-range dependency features by calculating global correlation through multi-head self-attention mechanism. In the STER[13] model, after the image is encoded, different scales of features are extracted through four Transformer Blocks downsampling, and then different-scale features are upsampled to the original resolution size. The final different-scale target is captured by superimposing the four feature maps in the channel dimension. The Segmentor[14], as a pure Transformer encoding-decoding architecture, utilizes the global image context of each layer in the model. In the original ViT model, the Mask Transformer is proposed to decode the encoder and class embedding output in the decoding phase, and the Argmax is applied after upsampling to classify each pixel and output the final pixel segmentation map. This further improves the performance of ViT on small training sets.

**Feature fusion**. Given the distinct and complementary properties of convolution and self-attention, there is potential for benefiting from both paradigms by integrating these modules. U-Net[15] employs skip connections to superimpose features from different levels, FPN connects multi-scale features through a top-down propagation path. ASPP in DeepLabv3[16] sets up dilated convolutions with different dilation rates to obtain features of different scales. In response to the limitations of the receptive field of convolutional neural networks, many studies consider fusing global and local features. SegFormer[17] uses a lightweight MLP decoder to aggregate information from different layers, thus combining local attention and global attention to present powerful representations. DS-Net simultaneously computes fine-grained and integrated features, proposes an intra-scale propagation module to handle the two different resolutions in each block, and proposes a cross-scale alignment module to perform inter-feature information interaction across two scales. Conformer[18] relies on Feature Coupling Units (FCU) to input local and global features separately into two parallel and concurrent feature processing flows, thereby completing feature interaction. The focus of this paper is on the effective interaction of global and local features to avoid interference from invalid information.

## III. METHOD

The proposed model (CrackFuse) is shown in Figure 1, based on a dual-stream architecture: concurrent feature extraction and processing are performed by the Transformer global processing stream and the convolutional local processing stream, and feature interaction and multi-level fusion are accomplished through the BiFuse Module interaction module. Finally, the edge optimization module is employed to enhance the accuracy of edge extraction by the model.

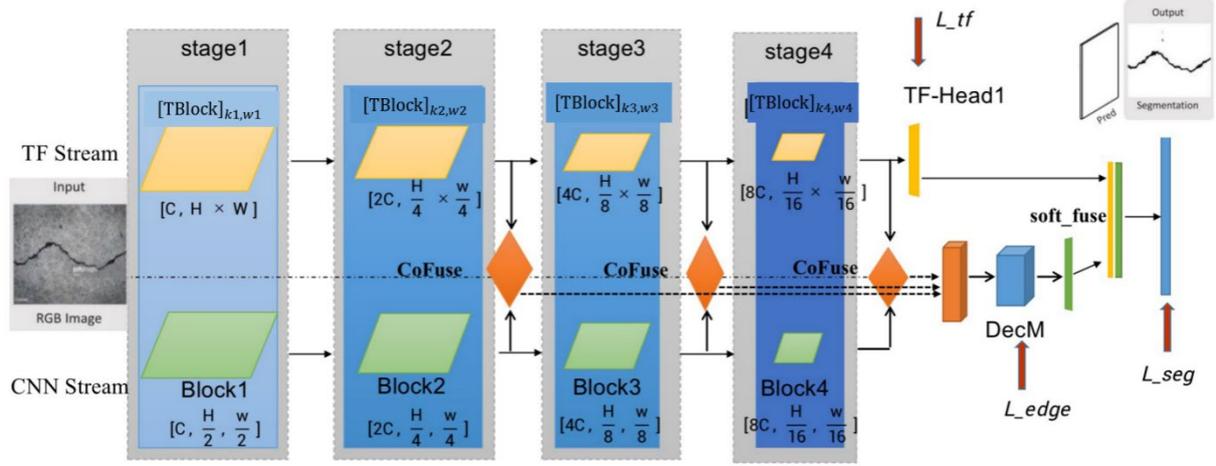

Figure 1 Overall structure diagram of the model

## 3.1 Dual-scale Representations(DSM)

CNN collects local features through convolutional operations and preserves local clues as features. Vision Transformer is considered to be capable of aggregating global representations in a soft manner through cascaded Self-Attention modules among compressed patch embeddings. By processing images in parallel, multi-level features of the image can be maximally preserved. For a given image $x$ with size of $W \times H \times 3$, the dual-stream network DSM is defined as D and the detection result M is obtained after $x$ is inputted into D. Let $x_i$ be a pixel in the image, $D(x_i)$ represents the predicted result for each pixel based on the dual-stream network, and $D(x_i) \in \{0, 1\}$. The set of predicted results for all pixels is the crack segmentation result of the dual-stream network D for image $x$, as shown in equation (1) below. The detailed structure of each module in network D is described in the following text.

$$M \leftarrow \{D(x_i)\}, \quad i = 1, 2, \ldots, W \times H \tag{1}$$

**Global representation** The lightweight Cswin Transformer model is used, and the image is divided into multiple patches. Each patch is encoded and inputted into the subsequent four blocks for computing long-range dependencies. Cross-attention is used, where each patch only focuses on the cross-shaped window portion located in the same row and column as itself. In the top-down computation, the attention window gradually expands layer by layer and finally aggregates the features of the entire image. Following the structure of the reference feature pyramid, multi-level features are better aggregated. After all layers of features are upsampled by interpolation to the same dimension, the final global feature of 4*channel is generated through skip connections. The overall formula for the global branch is shown in equations (2) and (3) below.

$$X_g^i = Transformer_{Block_i}(X_g^{i-1}) \quad (2)$$

$$X_g = F\left(Upsample(X_g^1 | X_g^2 | X_g^3 | X_g^4)\right) \quad (3)$$

The four-stage process of global processing branch is shown in Figure 2. The input of the first stage is the image block coding vector, the number of channels is C, and $\frac{H}{4} \times \frac{W}{4}$ is the total dimension of the vector. Convolution operation is used in each adjacent stage to reduce the feature scale and increase the number of channels. After the feature processing of stage i, the output feature graph size is $\frac{H}{2^{i+1}} \times \frac{W}{2^{i+1}} \times 2^i C$. Each $stage_i$ consists of $N_i$ modules $TBlock_i$, where $N_i$ is the parameter of the model.

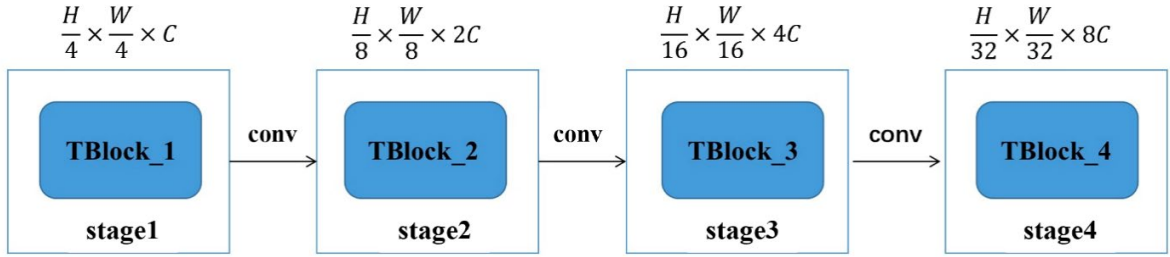

Figure 2 Four-stage processing of CSwin Transformer

The TBlock consists of the multi-head self-attention computing module Cross MSA and the forward propagation FFM module, with jump connections added to prevent the gradient from disappearing. Formula 4 and 5 are calculated as follows:

$$X_l = Attention(LN(X_{l-1})) + X_{l-1} \quad (4)$$

$$\widetilde{X}_l = MLP(LN(X_l)) + X_l \quad (5)$$

Where represents the output of the input TBlock module and is the output of the input TBLock module. $X_{l-1}$ represents the layer normalization processing, and MLP is the multilayer perceptron.

**Local representation** is to obtain more shallow information, such as details of edges and contours. Drawing on the edge detection model HED and based on the FPN network model, a bottom-up and top-down processing flow is designed. In the middle of the two reverse processing flows, phased features are fused to obtain more accurate features.

In the local processing branch, as shown in Figure 3 below, the fourth stage outputs the feature vector with the size of $B \times 512 \times 24 \times 24$. After three times of double upsampling, the feature scale gradually recovers to $B \times 256 \times 48 \times 48$, $B \times 128 \times 96 \times 96$ and $B \times 64 \times 192 \times 192$, where B is the number of images trained in each batch of the model. The result of down-sampling at each stage in the local processing branch is called $F_{l\_down}^i$, i representing the i-th stage, by the same token, the upsampled result is called $F^i_{l\_up}$. Finally, the sum of the same stage $F^i_{l_{down}}$, $F^i_{l\_up}$ is taken as the output of the local processing stream. Formula 6 is as follows:

$$F^i_l = f\left(F^i_{l_{down}} \big| F^i_{l_{up}}\right), \quad i = 2, 3, 4 \tag{6}$$

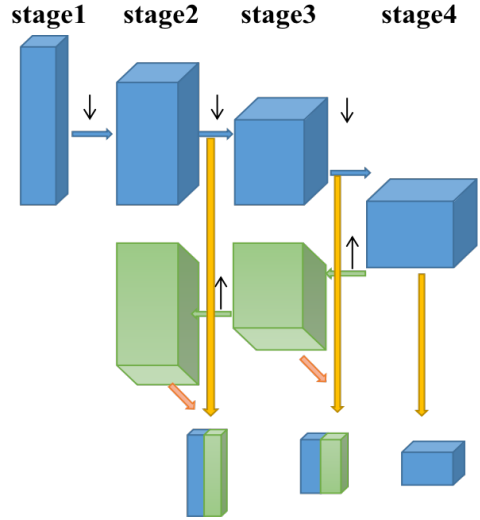

Figure 3 Local processing branch structure diagram

## 3.2 Fuse interaction Moudle(CoFuse)

**Feature Alignment** inserts parallel fusion modules in the middle of two processing streams for feature interactive fusion. The input of the module is the intermediate feature of the output of each stage in the global processing flow and the local processing flow, achieving the effect of simultaneous extraction and fusion. The features output by the global processing stream need to be reshaped into the same dimension as the local processing stream by a feature alignment operation.

**Feature Refine** Further refines and filters the input features in order to retain valid information and avoid interference from invalid information in the fusion process. Channel Attention draws on SENet[20] and correct the features of channels by re-modeling the relationship between channels through squeeze and extend operations, so as to improve the characterization ability of neural networks. Global information can be used to strengthen useful features and dilute useless ones. Firstly squeeze to compress the features of each channel as the descriptor for that channel, by mean pooling averaging the features within the channel. The specific operation formula (7-9) is as follows:

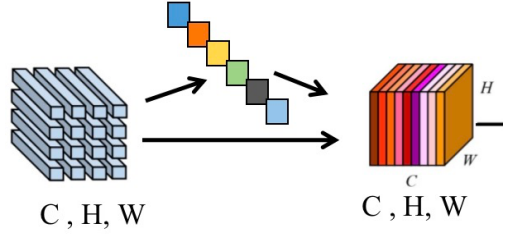

**Figure 4 Schematic diagram of channel attention calculation**

$$Z_c = f_{sq}(F_g) = avg(F_g) = \frac{1}{H \times W} \sum_{i=1}^{H} \sum_{j=1}^{W} F(i,j) \tag{7}$$

$$S = f_{ex}(\{Z_c \mid W\}) = \sigma(g(Z,W)) = \sigma(W_2 \cdot \delta(W_1 \cdot Z)) \tag{8}$$

$$\widetilde{F}_g = f(F_g, S) = F_g \cdot S \tag{9}$$

S is the vector of C dimension, is the weight of the learned channel, useless features will be tended to 0. Activate the function and $\delta, \sigma$ are selected as ReLU and sigmoid in turn. W is the set of weight $W_1$ and $W_2$ matrices, including Where the dimension is $\mathbb{R}^{\frac{C}{r} \times C}$ and $\mathbb{R}^{C \times \frac{C}{r}}$.

**Spatial Attention** Local features contain a lot of low-level details. But not all the low-level information is useful: background texture, lighting and other factors will often interfere with the model. We believe that channel attention can tell the model which features are meaningful, so we use spatial attention in local features to further remove these redundant noises.

$$M_s(F_l^i) = \sigma\left(f^{7\times7}([AvgPool(F_l^i), MaxPool(F_l^i)])\right) \tag{10}$$

The local feature is of $R^{H \times W \times C}$. First, maximum pooling and average pooling of channel dimensions are carried out to obtain 2 channel descriptions of {H×W×1}, and the two descriptions are splicing together according to channels. The weight coefficient is obtained after compress and sigmoid function activation. $M_s^i$ Multiply the input feature to get the scaled new feature.

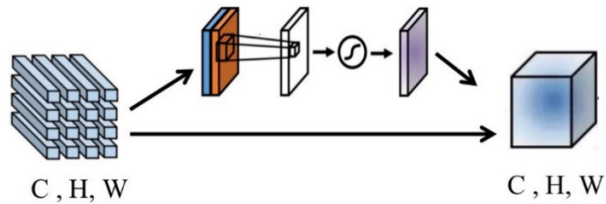

**Figure 5 Schematic diagram of spatial attention calculation**

$$F_l^i = F_{scale}(F_l^i, M_s^i) = F_l^i \cdot M_s^i \tag{11}$$

**Feature Interaction** Firstly, the combined features $F_{cat}$ consists of global features $X_g^i$

and local features $F_l^i$ are input into CoFusion module for feature interaction and correlation calculation. It consists of a MLP, using the softmax in front of the output of the MLP to generate adaptive weights for different channels. Finally, the obtained correlation probability graph is applied to the original feature graph to complete the feature interaction.

$$CorrMap^i = \partial\left(b_2 + W_2\left(b_1 + W_1([F_l^i, X_g^i])\right)\right) \tag{12}$$

$$F_{fuse}^i = CorrMap^i \times [F_l^i, X_g^i] \tag{13}$$

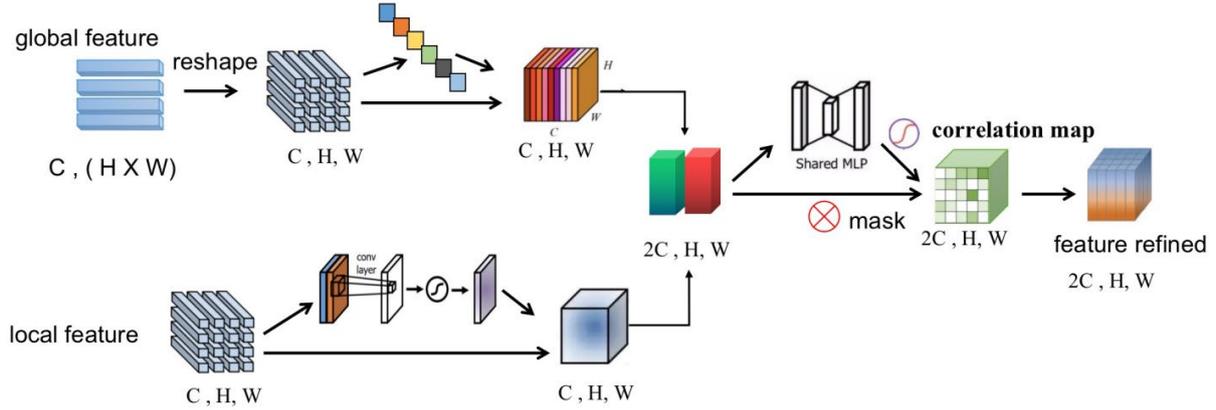

**Figure 6 Detailed structure diagram of feature fusion module**

### 3.3 Edge optimization(DecM)

In order to extract the edge information to the crack damage accurately we designed the edge optimization module. The main part of the feature is decoupled from the contour part and the edge supervision is added to optimize the feature extraction of the edge part. We use continuous down sampling to obtain the pseudo-clustering center, then the original feature is projected to correct according to the pseudo-clustering center. The edge information $F_{edge}$ can be obtained by subtracting the main feature from the original feature.

$$F_{flow} = \gamma\left([Up(Down(F_{fuse})), F_{fuse}]\right) \tag{14}$$

$$F_{body} = Grid_{sample(F_{fuse}, Grid^{h \times w} + F_{flow})} \tag{15}$$

$$F_{edge} = cat(F_{fuse} - F_{seg\_wrap}, F_{fuse}^1) \tag{16}$$

$$F_{final} = F_{edge} + F_{body} \tag{17}$$

### 3.4 Objective Function.

The losses adopted in this paper are divided into structural loss and edge loss. The structural loss is used to supervise the final prediction accuracy, the edge loss is used to supervise the accuracy of edge extraction. The overall loss is composed of the following parts:

$$L_{all} = \theta_0 L_{final} + \theta_1 L_{global} + \theta_2 L_{local} + \theta_3 L_{edge} \tag{18}$$

Where $\theta_0$, $\theta_1$, $\theta_2$, $\theta_3$ are the weight parameters, which are used to weigh the proportion of different losses.

Where $L_{final}$, $L_{global}$, $L_{local}$ is used to calculate the loss of the overall structure of the image. The overall loss is divided into two parts, including pixel level binary cross entropy loss and intersection ratio loss. Where P is the total number of all pixels, i is the certain pixel, and the true pixel value $y_i$, $P(x_i)$ is the predicted probability.

$$L_{BCE} = -\left(\sum_i^p y_i(logP(x_i) + (1 - y_i)logP(1 - x_i))\right) \quad (18)$$

$$L_{IOU} = \frac{\sum_i^P y_i\, P(x_i)}{\sum_i^P y_i + \sum_i^P P(x_i)} \quad (19)$$

In order to balance the large gap between the ratio of positive and negative pixels and weaken the influence of pixels that are difficult to identify to ensure the accuracy of the overall effect, the weight with weight $\omega$ is introduced as the loss weight.

$$L_{final} = \omega(L_{BCE} + L_{IOU}) \quad (20)$$

Taking into account the extremely unbalanced ratio of positive and negative pixels in a single image: often the edge pixels only occupy a small proportion. Therefore, based on the binary cross entropy at the pixel level, we introduce the adaptive weight mask as the pixel coefficient to calculate the edge loss.

$$L_{edge} = -\left(\frac{\sum(y_i = 0)}{N}\sum_i^p y_i\left(logP(x_i) + \frac{\sum(y_i = 1)}{N}(1 - y_i)logP(1 - x_i)\right)\right) \quad (21)$$

### IV. EXPERIMENTS

**Dataset.** This work makes use of four datasets, CrackForest[21], DeepCrack, CRACK500, CrackTree260[22].

- The CrackForest dataset(CFD) consists of a total of 118 images of size 480×320 pixels. The images have been taken from road surfaces in Beijing. This dataset is then split into 71 training, and 46 testing images.
- CrackTree260 (CT260) is a dataset containing 260 grayscale road pavement images of different sizes (800 × 600 and 960 × 720 pixels).
- A total of 537 RGB images are contained in Deepcrack. The dataset split is given as 300 training, and 237 testing images, all of size 544 ×384 pixels.
- CRACK500 is a dataset containing 3000 RGB images of size 800×600 pixels. This dataset is then split into 1500 training, 200 vailding and 1300 testing images.

**Evaluation Metrics.** We use F1 score and IOU to measure the quality of the generated images at pixel level.

**Implementation Details.** We use Adam optimizer with {β1 = 0.5, β2 = 0.999} and train the model for 50 epochs. The learning rate is set to 1e−4, and the batch size is 2.

**Comparison with SOTAs(State Of Arts).** We further conduct a comparison experiment with most related SOTA methods on the DeepCrack dataset. A total of six common concrete superficial crack segmentation methods and the classic semantic segmentation network model are selected. The following is the introduction of these models.

**U-Net** classical semantic segmentation model, U-Net model for encoder -decoder structure, it as a general network model, in the semantic segmentation of various tasks are very good performance.

**DeepLabv3+** uses the Xception module for task splitting and deep separable convolution into void convolution and decoder modules to achieve high accuracy in the model.

**R2UNet [23]** introduces residual blocks as well as loop structures on the basis of U-Net;

**DeepCrack** model is based on edge detection network HED, and a bypass branch is designed to enhance the edge information extracted from shallow network.

**STRNet[24]** model proposes a new high performance network of depth encoder and attention decoder, which uses channel attention to enhance features during feature extraction, and Transformer is used to build the decoder, adding separable convolution operations in order to achieve real-time detection rate. A semantically trainable representation network is proposed to improve.

**CrackFormer[25]** model uses attention mechanism to improve the model's crack detection in complex scenes, and proposes a novel attention mode Self AB. The encoder decoder structure based on Transformer integrates Self AB module and Scale AB module. Where Self AB is embedded into different levels of encoder and decoder modules and Scale AB introduces corresponding decoder between encoder feature map and encoder feature map.

The comparison model is selected based on two principles: first, the source code is disclosed and the code is reproducible. second, the paper claim that the model has achieved the experimental effect of SOTA. Quantitative experiment and qualitative experiment were carried out respectively in the comparative experiment. Quantitative experiment refers to the numerical comparison of F1 and IOU on the public data set, as shown in the table below. Qualitative experiment refers to the visualization results of crack damage segmentation on superficial concrete images, as shown in the figure 8. All experimental results are reproduced in the software environment required in the original paper, and the software used is Nvidia 3080 GPU, and the parameters used in the model are consistent. In this section, the CRACK500 and DeepCrack datasets were selected as the datasets for the crack segmentation task. This is because these two data sets contain the largest number of images, and the images involve rich scenes, which can detect the accuracy of model segmentation and its generalization performance at the same time.

**Table 1 Comparison table of experimental results between Dec_DSFM model and existing model**

| Model | F1 | | IOU | | FLOPs(G) | Params(M) |
|---|---|---|---|---|---|---|
| | CRACK500 | DeepCrack | CRACK500 | DeepCrack | | |
| U-Net | 69.4% | 90.6% | 58.0% | 83.2% | 70.02 | 13.40 |
| DeepLabv3+ | 67.3% | 89.5% | 55.9% | 81.5% | 430.36 | 137.67 |
| R2UNet | 45.1% | 86.7% | 34.1% | 77.4% | 344.06 | 39.09 |
| DeepCrack | 64.8% | 90.5% | 58.3% | 82.7% | 307.81 | 30.91 |
| STRNet | 65.6% | 88.1% | 54.8% | 79.3% | 7.78 | 2.08 |
| CrackFormer | 69.8% | 89.9% | 58.6% | 82.1% | 46.04 | 2.08 |
| **CrackFuse** | **78.1%** | **93.7%** | **66.0%** | **86.6%** | **293.05** | **111.43** |

The optimal results in the table are shown in bold. As can be seen from the data in the table, the segmentation accuracy of CrackFuse achieve highest score. The segmentation results on the two public data sets are 6.0% and 6.0% higher than the average F1 value and the average IOU value of the CrackFormer with the best performance at present. Images in the CRACK500 dataset contain a variety of cracks: linear, crossed and reticulated. The background texture of the image is complex, in which the contrast between the crack area and the background area is low, which makes the crack segmentation more difficult. The CrackFuse has shown satisfactory results on this dataset, proving its versatility and strong generalization ability for complex scenes. It solves the problem that only a single scene fracture can be detected in some models. The number of images in DeepCrack data set is small, and the width of crack loss area is generally larger than that in CRACK500. It can be seen from the data in the table that the image segmentation accuracy on DeepCrack data set is higher. Among the existing models, U-Net model has a better effect on small data sets. The DeepCrack model adds edge extraction tributaries to further strengthen the accuracy of segmentation results through deep supervision. CrackFormer added Transformer layer to extract long distance dependencies, but the detection effect on small data sets is not good, because of its attention computing is fully connected structure, Transformer on some features of the language, such as sequential, syntax, etc., has no prior inductive bias, and Transformer often needs a lot of data input to learn inductive bias. On the basis of convolutional neural network, STRNet adds compression and expansion operations to correct the weight between channels, further strengthening the characterization of model features. However, the STRNet model does not perform well on the CRACK500 dataset due to the lack of long distance dependencies, thus demonstrating the powerful feature extraction capability of the Dec-DSFM model. Therefore, through comprehensive comparison with the six SOTA methods, the Dec-DSFM model proposed in this paper achieves the optimal effect on the two indexes of the two public data sets, and has remarkable generalization ability, advancement and superiority.

The following figure shows the segmentation results of a comparison experiment between six crack detection models and the model proposed in this paper. The first row is the original image to be detected, the second row is the annotation of the image, the third row is the segmentation results of the model proposed in this paper, the fourth to ninth rows are the segmentation results of the other six models, and the optimal segmentation results are marked with red dotted lines. It can be seen from the display results that the accuracy of the model segmentation is higher than that of other models. U-Net model and DeepLabv3+ model cannot identify fine crack damage, and R2UNet model is easy to be disturbed by noise, resulting in some noise islands on the segmentation results. DeepCrack model has poor segmentation results for slender cracks, and the lack of long-distance dependence leads to disconnection of long-distance cracks. Both STRNet and CrackFormer models are attention-computing models. STRNet model focuses on ensuring real-time detection rate but cannot take into account the accuracy of segmentation. As a result, the model is prone to interference from some local redundant information, such as contrast, etc. The CrackFormer model cannot accurately identify slender cracks, because the model is completely based on Transformer construction, which fully extracts the high-level semantic information while ignoring the low-level semantic information. Figure 7 shows the segmentation results F1 value and IOU value on the verification set of the above seven crack segmentation models in the training process as the number of training iterations changes, and the results with the highest accuracy are marked with red lines. As seen from the figure, the model proposed in this paper has been performing well in the training process, and all evaluation indexes are higher than other crack loss segmentation models. With the increase of the number of training iterations, the features learned by CrackFuse are gradually saturated, and the segmentation accuracy of the model is maintained at a high level.

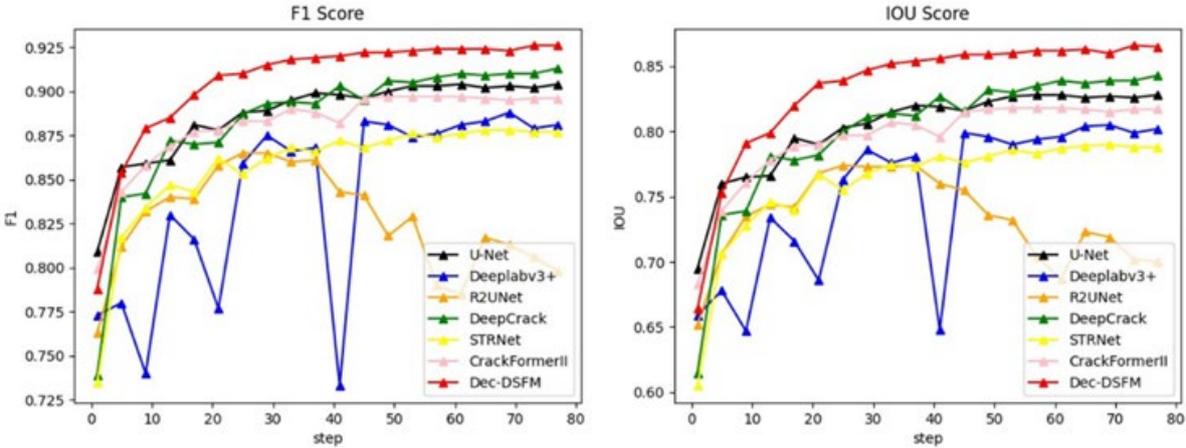

**Figure 7 Experimental results of CrackFuse and SOTAs**

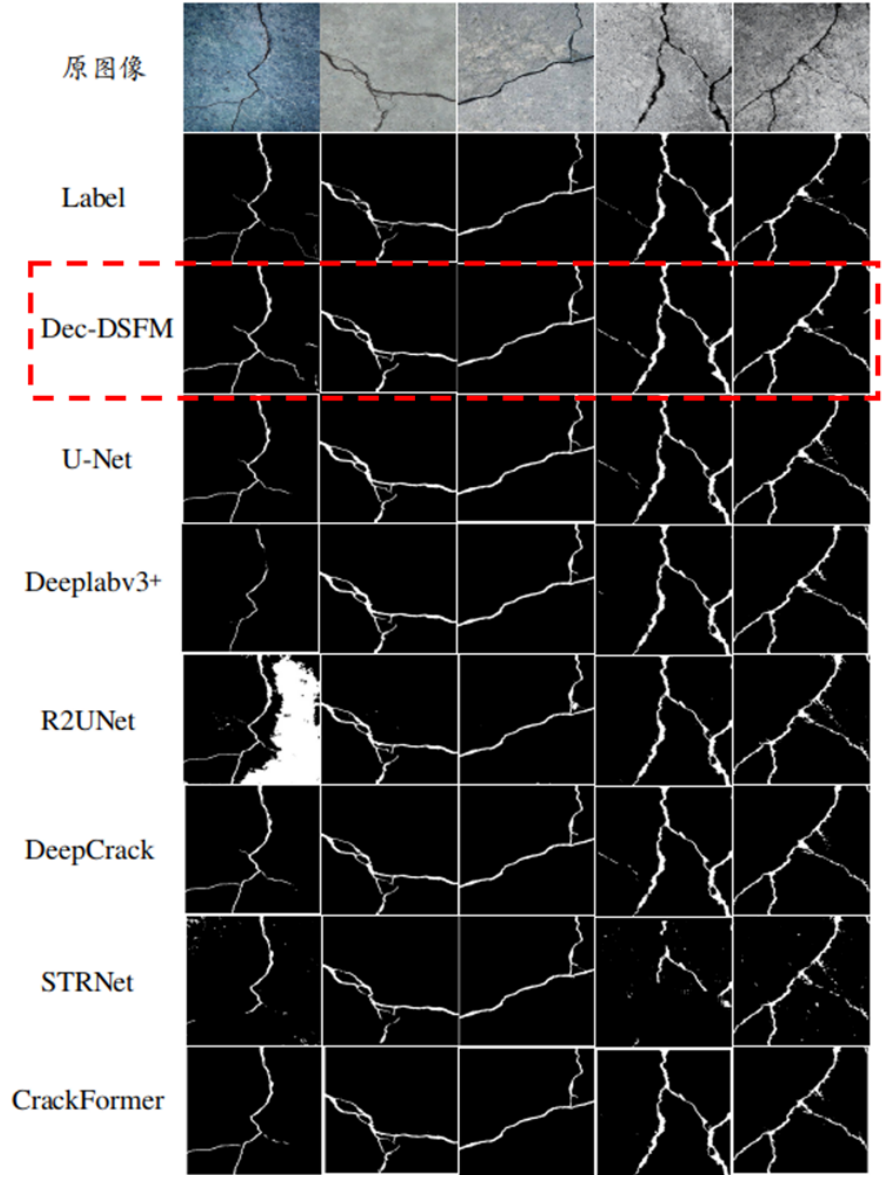

Figure 8 Segmentation results of CrackFuse and SOTAs

**Ablation experiments** were conducted on the CoFuse module to demonstrate its effectiveness. In order to more clearly demonstrate the effectiveness of each step in the design of this fusion module, in this experiment, the fusion module is divided into three parts: global feature filtering module, local feature filtering module, and correlation calculation module. In the ablation experiment, the effectiveness of these three modules will be evaluated. The experimental content includes the following five parts: baseline model (DSM), combination of dual flow network and global feature filtering module (DSM + gf_filter), combination of dual flow network and local feature filtering module (DSM + lf_filter), combination of dual flow network and correlation calculation module (DSM + corr_fuse) The combination of a dual flow network and two feature filtering modules (DSM+filter), and the combination of a dual flow network and a feature fusion module CoFuse (DSM+CoFuse). The specific experimental results on four datasets are shown in the table 2.

**Table 2 Ablation experiments F1 results of CoFuse**

| Model | F1 | | | | Average of F1 |
|---|---|---|---|---|---|
| | CRACK500 | DeepCrack | CrackTree260 | CrackForest | |
| DSM | 70.1% | 91.0% | 22.7% | 52.9% | 59.1% |
| DSM + gf_filter | 71.2% | 89.7% | 48.3% | 66.8% | 69.0% |
| DSM + lf_filter | 71.5% | 91.8% | 48.4% | 67.4% | 69.7% |
| DSM + corr_fuse | 71.6% | 92.5% | 48.1% | 67.9% | 70.0% |
| DSM + filter | 72.2% | 91.5% | 48.5% | 67.9% | 70.0% |
| **DSM + CoFuse** | **72.3%** | **92.6%** | **50.4%** | **68.5%** | **71.0%** |

**Table 2 Ablation experiments IOU results of CoFuse**

| Model | IOU | | | | Average of IOU |
|---|---|---|---|---|---|
| | CRACK500 | DeepCrack | CrackTree260 | CrackForest | |
| DSM | 64.6% | 83.5% | 11.6% | 34.1% | 48.4% |
| DSM + gf_filter | 59.8% | 84.7% | 33.5% | 51.5% | 57.3% |
| DSM + lf_filter | 60.1% | 85.2% | 33.6% | 51.5% | 57.6% |
| DSM + corr_fuse | 60.3% | 86.3% | 33.3% | 52.0% | 57.9% |
| DSM + filter | 60.9% | 84.7% | 33.6% | 52.0% | 57.8% |
| **DSM+ CoFuse** | **60.9%** | **86.5%** | **35.2%** | **52.5%** | **58.8%** |

**Table 4 The ablation experimental results F1 of the edge optimization(DecM)**

| Model | F1 | |
|---|---|---|
| | DeepCrack | CrackForest |
| DSFM | 92.6% | 68.6% |
| DSFM + DecM | 92.7% | 68.8% |
| DSFM + DecM + BCELoss | 92.2% | 68.5% |
| **DSFM + DecM + EdgeLoss** | **93.7%** | **69.3%** |

**Table 5 The ablation experimental results IOU of the edge optimization(DecM)**

| 模型/策略 | IOU 值 | |
|---|---|---|
| | DeepCrack | CrackForest |
| DSFM | 86.5% | 52.8% |
| DSFM + DecM | 86.7% | 53.1% |
| DSFM + DecM + BCELoss | 86.0% | 52.9% |
| **DSFM + DecM + EdgeLoss** | **88.3%** | **53.7%** |

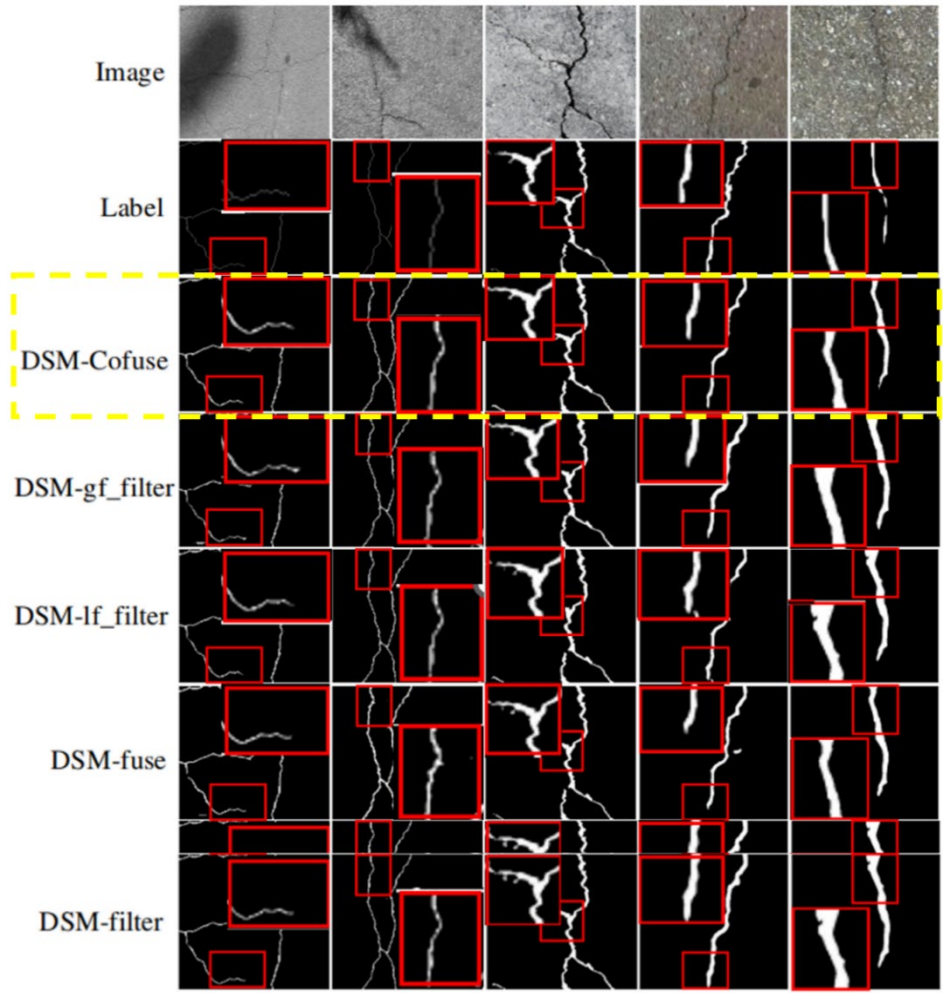

Figure 9 visualization of ablation experimental of CoFuse

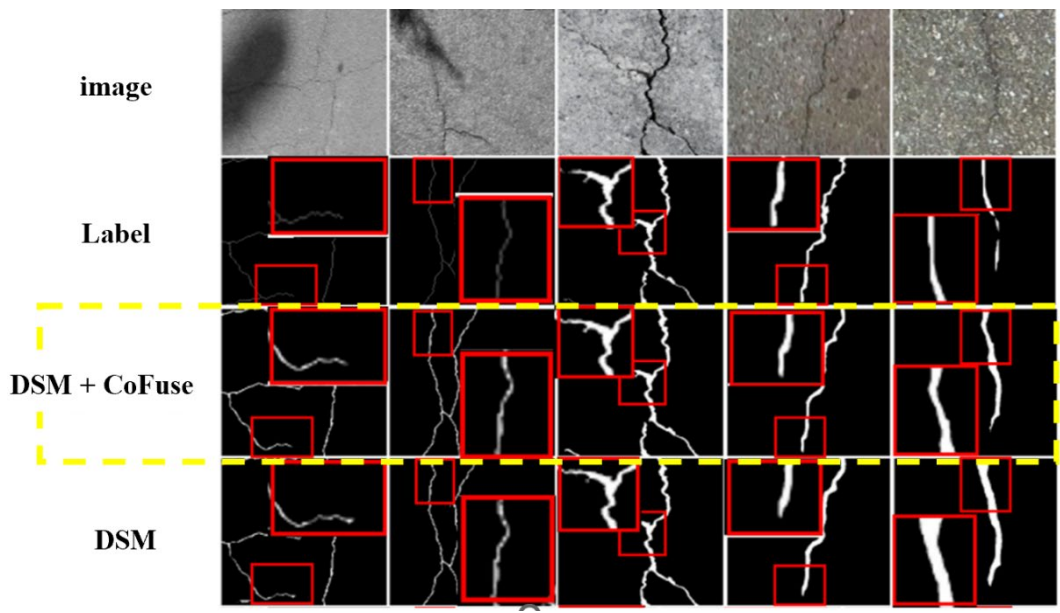

Figure 10 visualization of ablation experimental CoFuse

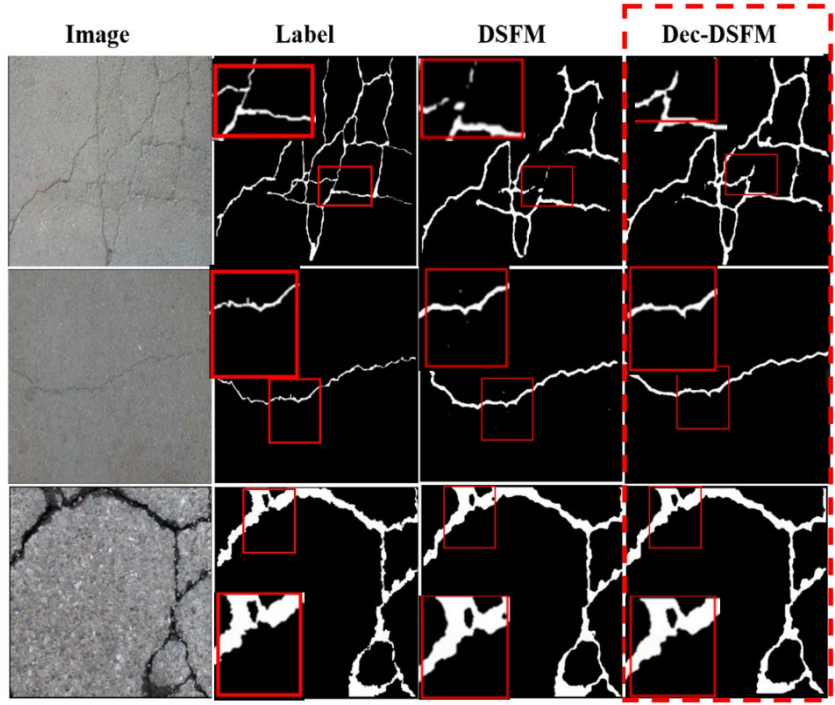

**Figure 10 visualization of ablation experimental of DecM**

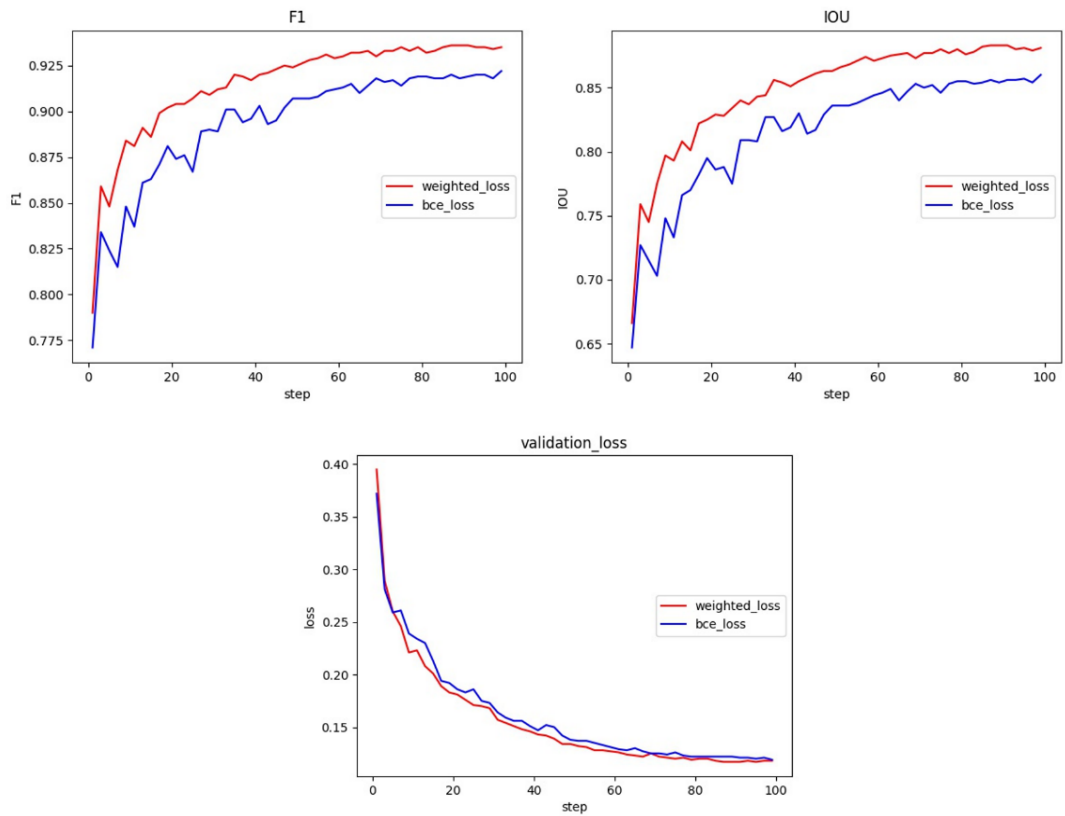

**Figure 11 Experimental results of different loss strategies**

## V. CONCLUSION

In this paper, A dual flow architecture is adopted. Transformer model is proposed to obtain the long-distance dependence relationship, which improves the generalization performance of the model and makes our model have better detection effect in the complex background of the real scene. On this basis, in order to fuse the global features and local features of the model, we put forward the dynamic feature fusion module, in the process of feature extraction feature interaction, so as to better use of the global and local features. In the end, the decoupling method is used to separate the main part and the edge part of the image, which effectively improves the segmentation accuracy of the edge. In the future, we will further optimize the extraction of edge parts, lightweight network model, and adapt to multi-scene crack detection and other defect detection.

## ACKNOWLEDGE

Thanks to Beihang Laboratory for support in this work, and to the teachers and classmates in the same laboratory for their support in my work.